\documentclass[conference]{IEEEtran}
\IEEEoverridecommandlockouts
\usepackage{cite}
\usepackage{amsmath,amssymb,amsfonts}
\usepackage{algorithmic}
\usepackage{graphicx}
\usepackage{textcomp}
\usepackage{xcolor}
\usepackage{url}
\usepackage{verbatim}
\usepackage{hyperref}
\usepackage{soul}
\usepackage{graphicx}
\usepackage{float}
\usepackage{multirow}
\usepackage{array}
\usepackage{fancyhdr}
\def\BibTeX{{\rm B\kern-.05em{\sc i\kern-.025em b}\kern-.08em
    T\kern-.1667em\lower.7ex\hbox{E}\kern-.125emX}}

\begin{document}


\title{
Foreign Object Debris Detection for Airport Pavement Images based on Self-supervised Localization and Vision Transformer

\thanks{
This paper has been accepted for publication by the 2022 International Conference on Computational Science \& Computational Intelligence (CSCI'22), Research Track on Signal \& Image Processing, Computer Vision \& Pattern Recognition.
}
}

\author{
    \IEEEauthorblockN{
    Travis Munyer\IEEEauthorrefmark{1},
    Daniel Brinkman\IEEEauthorrefmark{1}, 
    Xin Zhong\IEEEauthorrefmark{1}, 
    Chenyu Huang\IEEEauthorrefmark{2},
    Iason Konstantzos\IEEEauthorrefmark{3}}
    \smallskip
    \IEEEauthorblockA{\IEEEauthorrefmark{1}
    Department of Computer Science, University of Nebraska Omaha, Omaha, NE, USA
    \\\{tmunyer, dbrinkman, xzhong\}@unomaha.edu}
    \smallskip
    \IEEEauthorblockA{\IEEEauthorrefmark{2}
    Aviation Institute, University of Nebraska Omaha, Omaha, NE, USA
    \\chenyuhuang@unomaha.edu}
    \smallskip
    \IEEEauthorblockA{\IEEEauthorrefmark{3}
    Durham School of Architectural Engineering and Construction, University of Nebraska Lincoln, Lincoln, NE, USA
    \\iason.konstantzos@unl.edu}
}













\maketitle

\begin{abstract}
Supervised object detection methods provide subpar performance when applied to Foreign Object Debris (FOD) detection because FOD could be arbitrary objects according to the Federal Aviation Administration (FAA) specification.
Current supervised object detection algorithms require datasets that contain annotated examples of every to-be-detected object. 
While a large and expensive dataset could be developed to include common FOD examples, it is infeasible to collect all possible FOD examples in the dataset representation because of the open-ended nature of FOD. 
Limitations of the dataset could cause FOD detection systems driven by those supervised algorithms to miss certain FOD, which can become dangerous to airport operations. 
To this end, this paper presents a self-supervised FOD localization by learning to predict the runway images, which avoids the enumeration of FOD annotation examples. The localization method utilizes the Vision Transformer (ViT) to improve localization performance. The experiments show that the method successfully detects arbitrary FOD in real-world runway situations.
The paper also provides an extension to the localization result to perform classification; a feature that can be useful to downstream tasks.
To train the localization, this paper also presents a simple and realistic dataset creation framework that only collects clean runway images.
The training and testing data for this method are collected at a local airport using unmanned aircraft systems (UAS).
Additionally, the developed dataset is provided for public use and further studies.

\end{abstract}


\section{Introduction}\label{sec:intro}

The Federal Aviation Administration (FAA) defines "Any object, live or not, located in an inappropriate location in the airport environment that has the capacity to injure airport or air carrier personnel and damage aircraft" as Foreign Object Debris (FOD)~\cite{fod-management}. 
FOD is responsible for billions of dollars in damages to aircraft each year~\cite{faa-fact-sheet}. 
Additionally, accidents caused by FOD can lead to injury or death. 
A robust, automatic, and affordable method for FOD detection is crucial for the safety of flight operations at airports, especially given the size and complexity of airports will keep increasing. 

\begin{figure}[t]
    \centering
    \includegraphics[width=0.75\linewidth]{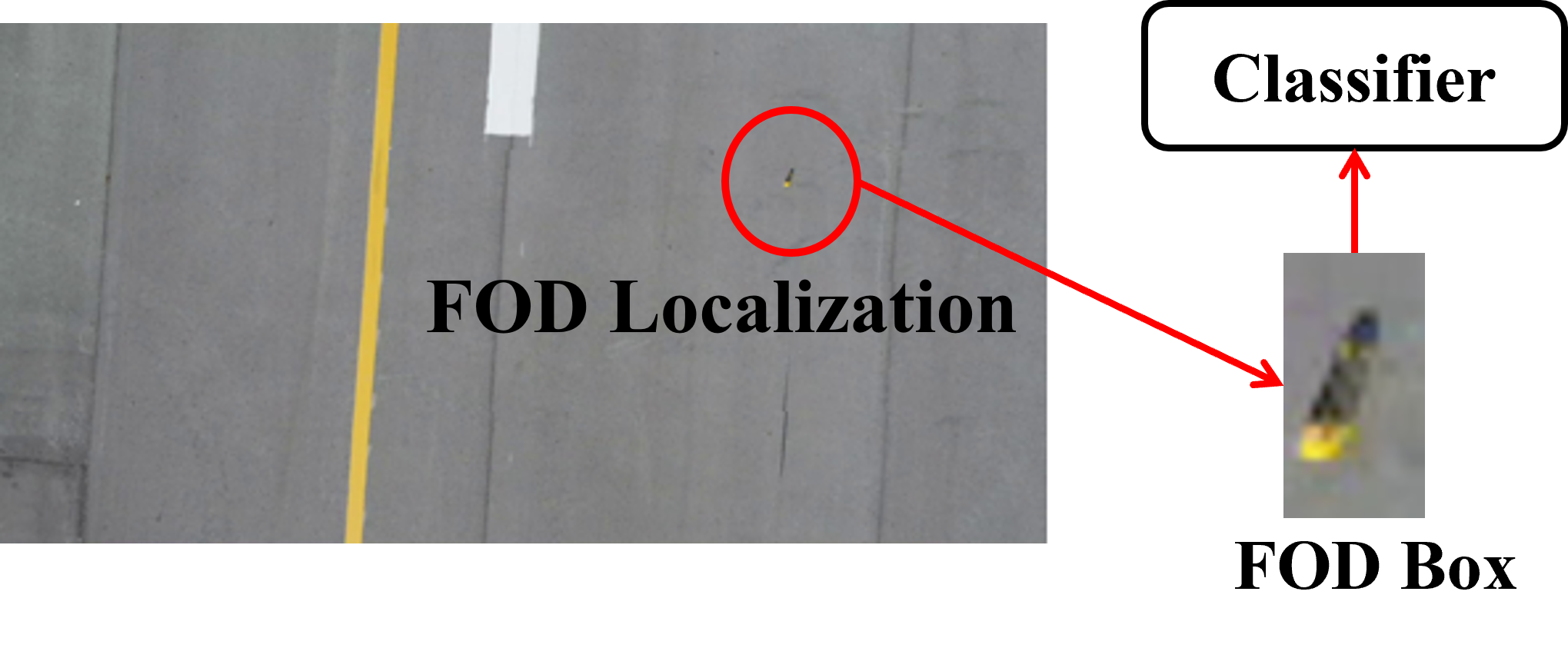}
    \vspace{-1.0em}
    \caption{The overview of the proposed FOD detection.}
    \vspace{-1.7em}
    \label{fig:fodsegmentation}
\end{figure}

Current FOD detection is primarily performed manually (\textit{i.e.}, FOD walks). 
Automated FOD detection methods can help reduce the negative impact of manual FOD detection on airport operations, and better address human error.
Most existing automated detection systems rely on radar-based technologies, however, those methods have not been widely adopted due to their high cost. 
For example, Boston Logan International Airport adopted one of those radar-based detection systems (\textit{i.e.,} FODetect) in 2013~\cite{faa-fact-sheet} for a total estimated cost was \$1.71 million~\cite{faa-fact-sheet}, which only included the installation of a single runway. 
More economical automated FOD detection techniques could be beneficial to large-scale prevention of costly aircraft safety events as more airports would be able to afford them. 
In addition, new detection techniques should be scalable to different airport environments and locations. 
Given the relatively low hardware cost to implement computer vision and deep learning-based methods (\textit{i.e.,} a camera at minimum) and the growing capabilities of Unmanned Aircraft Systems (UAS) technology, the integration of computer vision and UAS technology is expected to be advantageous over many existing FOD detection methods in airport operations. 
A novel FOD detection method based on this integrative approach is developed and introduced in this paper.  

There was a variety of computer vision-based FOD detection strategies proposed in previous studies, which will be discussed in more detail in section~\ref{sec:related}. One idea is to use supervised object detection algorithms such as YOLO and SSD~\cite{10.1145/3463677.3463743, YOLOV3forFOD}. Supervised detection methods are impractical for FOD detection because they can only detect predefined classes due to their dependence on a dataset with predefined classes and samples while FOD is a broad datatype. Another idea is to store a database of original runway and taxiway images and use direct image processing techniques to see if the images have changed. This requires excessive image collection that is specific to airports, and may not be robust to subtle changes in airport environments. A method that is capable of detecting previously unseen categories of objects and is robust to subtle changes in airport environments is more desirable and practical for FOD detection.



Recently, as an extension to the promising transformer~\cite{https://doi.org/10.48550/arxiv.1706.03762} in natural language processing, the Vision Transformer (ViT)~\cite{dosovitskiy2021an} has been developed in the area of deep learning and computer vision. 
Many breakthroughs~\cite{https://doi.org/10.48550/arxiv.2102.04306, 9578646, https://doi.org/10.48550/arxiv.2105.08582} have been enabled by the learning capacity of the ViT.
However, applying ViT in the design of the learning systems for FOD detection remains open. 
Hence, we explore the potential of ViT in the proposed method.

\begin{figure*}[!h]
    \centering
    \includegraphics[width=0.85\linewidth]{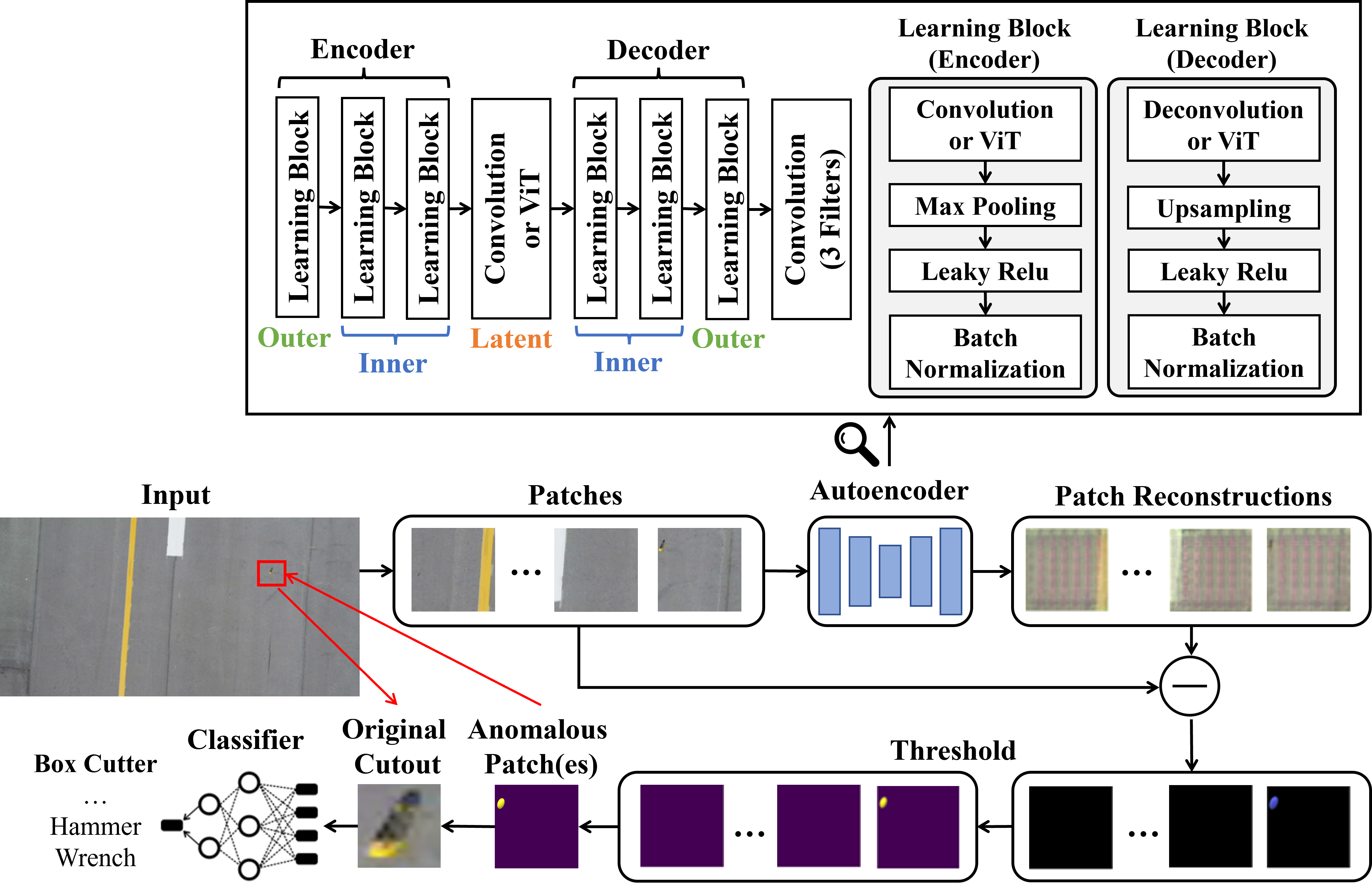}
    \vspace{-1.0em}
    \caption{Details of the proposed method: the localization, the network architectures, and the classification.}
    \vspace{-1.5em}
    \label{fig:methodfigure}
\end{figure*}

This paper provides a novel computer vision and deep learning-based FOD detection solution that is developed considering the limitations of existing methods and the practical demand of airport operators. 
The proposed strategy provides a FOD detection method that is economical to implement, not airport-environment-specific, and adaptive to previously unseen FOD. 
This self-supervised method adopts a runway image prediction/reconstruction mechanism to localize previously unseen objects and does not require human annotations.
Intuitively, the method is trained on a dataset of clean images, and anomalous regions provide the segmentation result at prediction time. 
This intuition is further explained in section~\ref{sec:construction}. The process is summarized in figure~\ref{fig:fodsegmentation} and is presented in more detail in figure~\ref{fig:methodfigure}.

The self-supervised localization process for FOD detection is the primary focus of this paper. 
In addition, we also extend this framework into an object detection method by cropping the segmented regions of the images, and feeding these cropped objects through a standard, pre-existing classification model. 
The inclusion of a supervised image classification component is a beneficial extension as the actual localization of FOD performed by the self-supervised portion of this network is a critical process. 
This is because airport operators may only be concerned with the actual discovery of arbitrary FOD objects and not necessarily the classification of these objects. 
The classification extension may facilitate downstream tasks, such as automated image collection for new FOD datasets (objects with a low classification score may be saved as new objects for future training data), filtering mislocalizations, automated removal of FOD (automatic removal may require object specific information such as average weight data associated with the object), and automated FOD logging/reporting system. 
The classification extension is also summarized in figure~\ref{fig:methodfigure}.

Another beneficial extension of the proposed self-supervised localization method is its data creation framework, which supports self-supervised FOD detection and performance evaluation for FOD localization. The framework allows the dataset to be extended efficiently, which is beneficial to downstream tasks.

In summary, the primary advantages of this work can be summarized as the following three contributions: 
(1) an extensible data collection framework is developed to support self-supervised FOD localization, 
(2) a self-supervised ViT-based FOD localization method is proposed, which is not dependent on annotated data, is not airport-specific, and is more economical than existing radar-based systems; 
and (3) an extension of the FOD localization to perform classification, which shows high accuracy of the proposed scheme. 
The rest of this paper is organized as the follows. 
We discuss related works in section~\ref{sec:related}. The dataset creation process and the FOD detection method is presented in more detail in section~\ref{sec:construction}. 
Experiment results are provided in section~\ref{sec:analysis}, and are followed by a conclusion and discussion of future work in section~\ref{sec:conclusion_discussion}.

\section{Related Work}\label{sec:related}

As we summarize our advantages as the dataset creation framework and the FOD detection method, we focus on the review of related work in these two categories. 
Section~\ref{subsec:RelatedDatasets} briefly reviews related datasets and section~\ref{subsec:FODExisting} revisits FOD detection method in the state-of-the-art. 

\vspace{-0.25em}
\subsection{Related FOD Datasets}
\label{subsec:RelatedDatasets}
A few FOD datasets were developed and published by previous studies, such as the dataset FOD-A~\cite{FOD-A}. However, this dataset is designed for the tasks of object detection or classification. All images contain FOD samples with bounding box annotation. As such, FOD-A is not directly usable for the localization method provided by this paper. The localization method requires a separated training/validation set which contains FOD-free images of runways, and a testing set which contains images that include FOD randomly distributed around the image. However, FOD-A is used for the classification extension provided in this paper.

\vspace{-0.25em}
\subsection{Existing FOD Detection Methods}\label{subsec:FODExisting}
Some examples of published FOD detection methods have attempted to use general object detection architectures (e.g. YOLO~\cite{redmon2018yolov3}, SSD~\cite{DBLP:journals/corr/LiuAESR15}), however, supervised object detection appears to be impractical for the FOD detection task~\cite{FOD-A, YOLOV3forFOD}. Any object that is improperly located in critical airport locations can be considered as FOD. It is not feasible to develop an image dataset that completely represents all possible types of FOD because of the possible broad range of FOD, which may prevent the common object detection methods from generalizing. Detection methods that cannot generalize could be unreliable for airport operations. 
Therefore, we conclude that supervised localization methods are not suitable. However, as long as the localization method can be generalizable, the classification can remain supervised. This is because the detection of FOD is the fundamental demand, while the classification extension is beneficial. 

\vspace{-0.2em}
Another method collects all clean runway images of an airport and stores them in an image database, then samples a new runway image at detection time, queries the image database for the corresponding image using GPS coordinates, aligns the two images, and then subtracts the two images to check for differences~\cite{FOD-Subtraction}. Areas of significant difference are a potential FOD detection. This type of method may not be robust to subtle changes in the airport environment. In addition, it requires the collection of images of all applicable airfield surfaces and such extensive image dataset may not be practical to collect and maintain for multiple airport implementations. Finally, it relies on the accuracy of GPS technology to find corresponding images, which may be error-prone. Inaccurate GPS estimates can cause detection failure if the wrong FOD-free image is used for comparison. Overall, this method could be fragile and may not easily scale to different airports. 
As discussed in more detail in section~\ref{sec:locAndClassification}, a new method is proposed to solve those key limitations that existing methods have. Specifically, this proposed method does not require the airport images to be stored for detection. The images are only required during training. In addition, the proposed localization method can generalize to previously unseen objects and is airport-independent.

\section{The Proposed Method}
\label{sec:construction}

This section describes the details of the proposed method.
Section~\ref{sec:data} discusses the data collection framework, Section~\ref{sec:locAndClassification} describes the FOD localization method, and Section~\ref{subsec:FODClassificationMethod} explains the classification extension.

\subsection{Data Collection Framework}\label{sec:data}

\begin{figure}[!b]
    \centering
    \vspace{-1.5em}
    \includegraphics[width=0.9\linewidth]{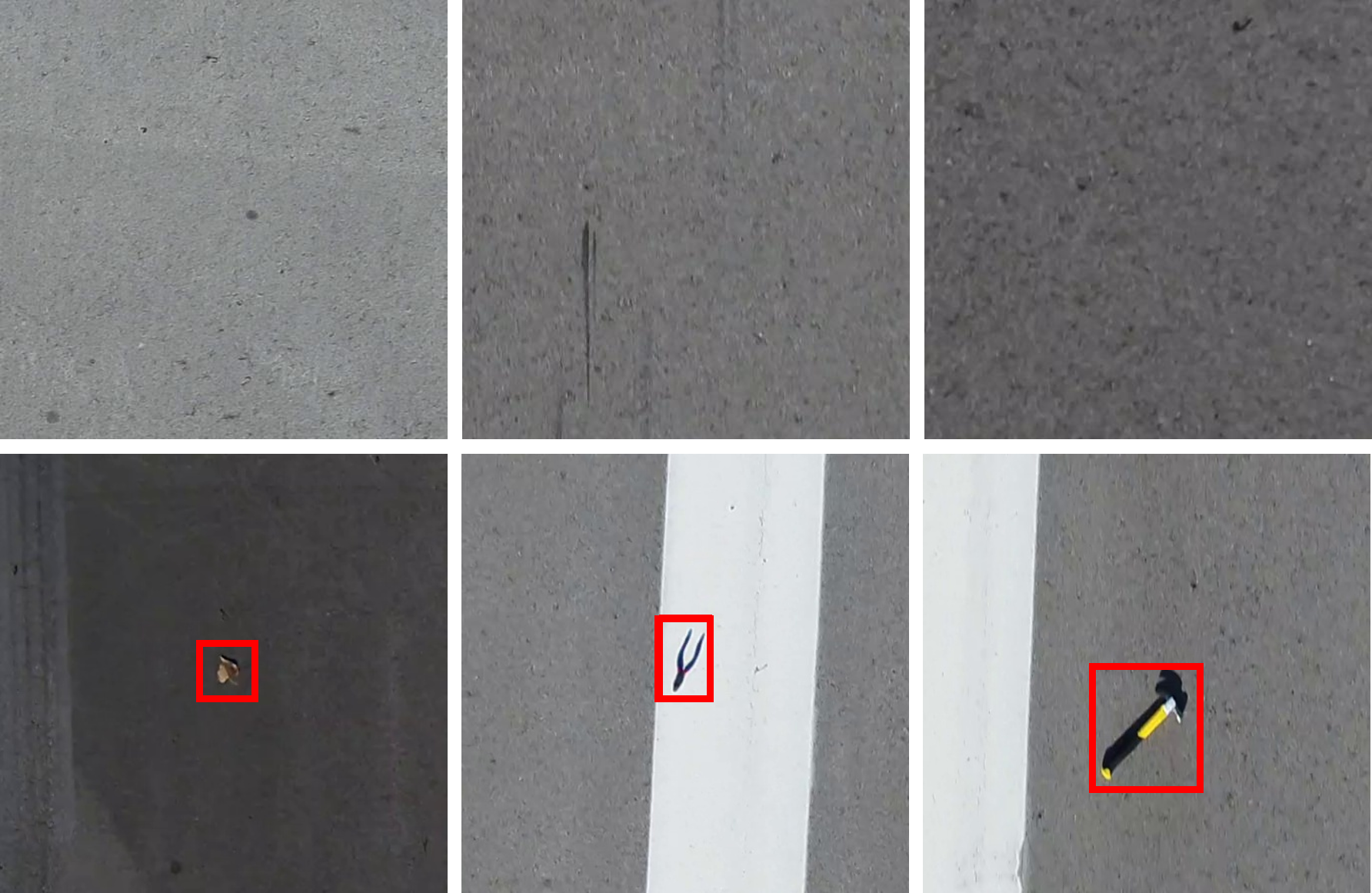}
    \vspace{-1.0em}
    \caption{Some examples of our collected data. Fist row: examples from the training set; Second row: examples from the testing set which have bounding boxes.}
    \vspace{-1.9em}
    \label{fig:dataexamples}
\end{figure}

\vspace{-0.25em}
The data is collected as videos from a local airport using UAS to reflect our goal of detecting FOD automatically from the aerial perspective. We collect the videos at $30$ feet, $60$ feet, and $140$ feet from the runway surface, providing ground sample distances of $0.1$ inch/pixel, $0.2$ inch/pixel, and $0.46$ inch/pixel respectively. After data collection, we found that the $60$ feet and $140$ feet videos lose too much detail, so $30$ feet videos are used in the dataset. The frame rate of videos is reduced to minimize duplicate frames, and the frames are separated to provide an image dataset. The $3840 \times 2160$ resolution frames were resized to the nearest multiple of $448 \times 448$ and then split into an $8$ by $4$ grid of $448 \times 448$ patches. This reduces the size of input images while retaining the detail of the collected data. The training dataset contains only clean images of runways and taxiways. The "clean" images do not contain FOD objects and therefore do not require annotation. The testing dataset contains videos of runways and taxiways where FOD objects were randomly distributed across the pavement. The testing dataset has bounding box annotation for FOD objects to support performance evaluation. Some examples of the training and testing data are shown in figure~\ref{fig:dataexamples}.

Using the described data creation framework, we are able to collect $81,185$ images for training efficiently. 
Processing the testing data results in $447$ testing patches. 
Each of these $447$ patches are annotated with bounding boxes for evaluation purposes. Within each $448 \times 448$ patch with FOD, the FOD object is manually annotated with a bounding box using the Computer Vision Annotation Tool (CVAT)~\cite{CVAT}. The annotations are then exported from CVAT and converted into a CSV file. These annotations are provided with the dataset.  

\subsection{FOD Localization}
\label{sec:locAndClassification}

The process of the method is as follows (see figure~\ref{fig:methodfigure} for a visual representation): the 3840x2160 resolution images, a resolution which is commonly considered to be high, are split into patches to preserve the detail of the images while reducing the computational burden. 
The proposed method provides FOD localization in the patches using a reconstruction technique. 
The reconstructed patches are used to propose patch-specific segmentation maps that label the background and the anomaly. 
As needed, the patch-specific segmentation maps can be combined to provide a full image segmentation or to display the FOD localizations on the entire image. 
Anomalous areas are cropped from the patch-specific segmentation map (the actual cropping is done on the original patch, the segmentation map provides the location), and normalized before classification.

In more detail: The reconstruction component of our method uses an autoencoder~\cite{doi:10.1126/science.1127647} with the architecture shown in figure~\ref{fig:methodfigure}. We organized the autoencoder structure into what we call learning blocks to facilitate experimentation with ViT layers.
A \textit{learning block} consists of the four layers presented in figure~\ref{fig:methodfigure}. The first layer within the block can be substituted with either a convolutional layer or a ViT layer~\cite{dosovitskiy2021an}. 
The ViT layer is an adaptation of the ViT classifier with the classification head removed. 
Most layers of the autoencoder are learning blocks, with the exception of the final layer, as depicted in figure~\ref{fig:methodfigure}. 
We consider the latent layer a learning block to simplify the definition, even though it contains only the convolutional or ViT layer. The locations of each learning block, as shown in the same figure, are labeled as an \textit{outer} block, an \textit{inner} block, a \textit{latent} block, an \textit{encoder} block, and/or a \textit{decoder} block. Each learning block is assumed to be used in its default configuration, where the first layer is convolutional. The learning blocks with ViT layers are mentioned explicitly with the location where the ViT learning block occurred in the autoencoder. For example, a latent ViT autoencoder implies the learning block in the latent location has a ViT layer, and the rest of the learning blocks are convolutional. 

To train the autoencoder for FOD localization, we minimize the following mean squared error (MSE) loss:

\begin{equation}
    \mathcal{L}_{loc} = \frac{1}{N}\sum(\hat{Z} - Z)^{2},
    \label{eq:loc_loss}
\end{equation}

\noindent where $Z$ is a clean image and $\hat{Z}$ is its model reconstruction. The clean images contain a runway/taxiway background and do not contain FOD, as discussed in section~\ref{sec:data}. Intuitively, the autoencoder learns to reconstruct the clean images, and fails to reconstruct portions of the images that contain new objects, as long as the autoencoder structure is chosen carefully. This selection process is detailed in section~\ref{sec:analysis}.

We employ this trained autoencoder to detect FOD as follows: First, the collected images have a high resolution as discussed previously. Utilizing high-resolution images is beneficial to preserving the detail of very small FOD. We found that directly resizing the high-resolution images prior to localization effectively deleted the presence of objects by modifying the image in a way similar to zooming out. To address that, we take the 3840x2160 resolution images and split them into patches with a width $N$ and a height $M$. This provides sub-images, or patches, of size $N \times M$. Instead of zooming out as the direct resize does, the patching effectively zooms into the patch location which improves small-object detection, an important characteristic for FOD detection. This benefit is observed empirically.

After the patching, the next step is the reconstruction. We reconstruct the patch $P$ into patch $P'$ by inputting $P$ into the trained autoencoder. Then, the difference matrix $D$ is calculated as the absolute difference between $P'$ and $P$: 
\begin{equation}
    D = |P' - P|.
    \label{eq:recom_absdiff}
\end{equation}

\noindent $D$ can also be calculated using the structural similarity calculation, which provides slightly different results. We chose the absolute difference method empirically. Values in $D$ that are close to 0 mark areas of the images are similar, and therefore not anomalous. Values in $D$ that are closer to the maximum pixel value (maximum pixel value is usually 1 or 255 depending on scaling) mark areas that are anomalous. This is because areas that remain similar after reconstruction will have similar pixel values, while areas with large reconstruction error will have a large difference between the pixel values. Then, we threshold $D$ to produce segmentation map $S$ using a threshold value given by Otsu's method. An example of the localization process is given in figure~\ref{fig:methodfigure}.

\subsection{FOD Classification}\label{subsec:FODClassificationMethod}
\begin{figure}[t]
    \centering
    \includegraphics[width=0.25\linewidth]{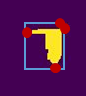}
    \vspace{-1.0em}
    \caption{An example bounding box computed from the extreme points.}
    \vspace{-1.7em}
    \label{fig:extremepoints}
\end{figure}

To convert the segmentation localization $S$ into the bounding box localization $R$, which is used for classification and evaluation, we calculate the extreme points on the segmentation map. The extreme points of the segmentation map is the segmented point the furthest left, the segmented point the furthest right, the segmented point closest to the top of the segmentation map, and the segmented point closest to the bottom of the segmentation map. From here, the four coordinates of a bounding box are computed directly from the extreme points to produce the bounding box localization $R$. An example of producing a bounding box from the extreme points is provided in figure~\ref{fig:extremepoints}.

We then crop $P$ using $R$ to produce the cropped localization $C$. From here, the method uses a mainstream supervised classification architecture, which is chosen empirically. We crop all the images from the FOD-A dataset~\cite{FOD-A} at the bounding boxes to create classification scenarios similar to the localization result. The cropped version of FOD-A is used to train the classification model minimizing the following categorical cross-entropy loss:

\begin{equation}
    \mathcal{L}_{class} = -\sum_{i=0}^{n-1}y_{i}\cdot\log\hat{y_{i}},
    \label{eq:class_loss}
\end{equation}

\noindent where $n$ is the number of classes, $y_{i}$ is the classification label, and $\hat{y_{i}}$ is a classification output from the model. 
Once the classification model is trained, we define $C$ as classified using the classifier.
The classification enables downstream tasks. For example, if the classification of $C$ results in a low prediction score below some threshold, $C$ can be labeled as unknown and saved for later manual labeling as it is likely not an image contained in the classification dataset. Otherwise, if the prediction score is above the chosen threshold, $C$ is classified accordingly. 

\section{Evaluation}\label{sec:analysis}

This section presents the experimental analysis of the proposed scheme. 
The evaluation method is presented in Section~\ref{subsec:evalmethod}, which is followed by a discussion of the localization experiments in Section~\ref{subsec:localizationexperiments}. 
Then, the results of the classification experiments are given in Section~\ref{subsec:classificationexperiments}. 
These subsections are followed by a comparison with previous related work in Section~\ref{sec:relatedcomparison}.

\subsection{Evaluation Method}\label{subsec:evalmethod}
The metric used to evaluate the localization results is the detection rate. Since the evaluation data is annotated as bounding boxes, and we also include a method to crop the result for optional classification, we use Intersection over Union (IoU) to determine to accuracy of a single localization. If the IOU of a proposal and the ground truth are above a set threshold, the localization is considered correct. Therefore, detection rate is defined as the number of correct localization, divided by the total number of expected localization.

\subsection{Localization}\label{subsec:localizationexperiments}
The strength of the autoencoder must be tuned carefully. We define "weak" and "strong" autoencoders to simplify the discussion of results. If the autoencoder is strong, it does not provide enough reconstruction error to segment the image. A strong autoencoder does not provide useful localizations because the reconstructions are too similar to the original image. Likewise, we define an autoencoder to be weak if there is too much reconstruction error for image segmentation. 
The weak autoencoder provides a reconstruction that is too dissimilar from the original image to provide an interesting localization. This means that a very low reconstruction error does not necessarily correspond to a quantifiable localization result in final testing. An ablation study format, using the detection rate metric for comparison, is used to determine a high performing autoencoder architecture. We experiment with different autoencoder depths and ViT locations. The effects of including skip connections, such as with a U-Net architecture, are also tested. 

The depth, or the number of learning blocks in the encoder and decoder is the first parameter we experiment with. A larger depth reduces the size of the latent space. We first start with a depth of two, which is the autoencoder with two learning blocks, each in the encoder and decoder. The autoencoder with a depth of two was too strong, and did not provide final segmentation maps using the method. We omit these results from table~\ref{tbl:ae} for this reason. Next, a depth of three is used. This autoencoder reconstructed images well, therefore this depth was used in the final experiments to compare between models further in table~\ref{tbl:ae}. We also tried a depth of four, but this resulted in an extremely small latent space and was therefore too weak. The autoencoder with a depth of four is also ommited from the table, as it did not provide interesting localization results. In these experiments, the patch size was $448 \times 448$. A larger patch size may increase the required depth, and a smaller patch size may decrease the required depth.

With the depth of the model determined and fixed to three, we next experiment with the impact of ViT layers at various locations of the model. First, a baseline model consisting of only convolutional layers is trained. This model performs well for localizing FOD as shown by the detection rate result in table~\ref{tbl:ae}. Next, the effect of including a ViT layer at the latent block is tested. This caused an overly weak autoencoder that did not provide a quantifiable detection rate. The next model is the Outer ViT model. This model also localized FOD well, as shown by the detection rate in table~\ref{tbl:ae}. 

Finally, we tried using a convolutional architecture with skip connections, similar to U-Net~\cite{https://doi.org/10.48550/arxiv.1505.04597}, as an additional baseline. The inclusion of skip connections resulted in a overly strong autoencoder, which can be determined by the skip-connection architecture reconstruction MSE and lack of localization results as given in table~\ref{tbl:ae}. 

\begin{figure}[h]
    \centering
    \vspace{-0.4cm}
    \includegraphics[width=0.7\linewidth]{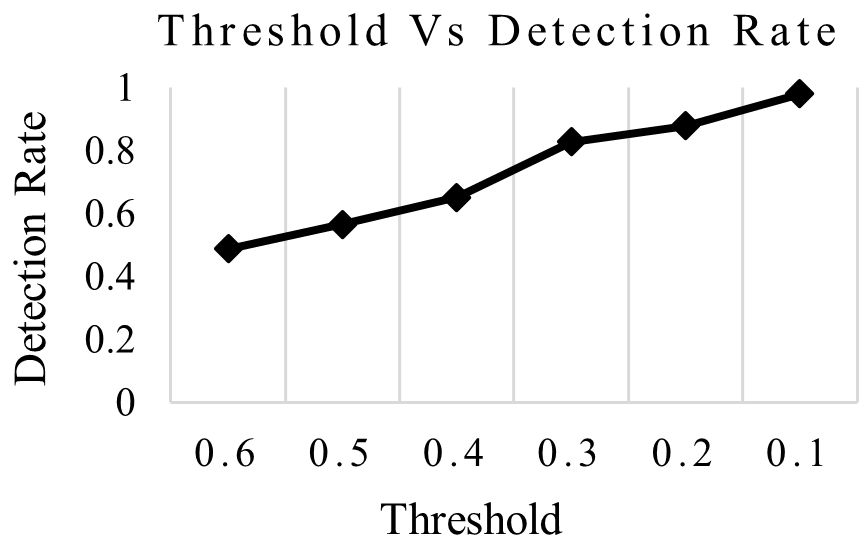}
    \vspace{-1.0em}
    \caption{Threshold v.s. Detection Rate.}
    \vspace{-0.75em}
    \label{fig:threshold}
\end{figure}

An IoU result above 0.3 is considered a confident localization when considering human error in dataset labeling, therefore 0.3 is the threshold used in the evaluation. The plot of threshold vs detection rate values for our method is given in figure~\ref{fig:threshold}, which shows the increase in detection rate for the outer ViT architecture as the chosen threshold value is reduced. The largest threshold with the most substantial increase in detection rate, this value being 0.3, is chosen as the threshold for comparison.
And as shown in table~\ref{tbl:ae}, the highest performing architecture is the Outer ViT autoencoder, which will be used for final comparison with related methods.

\vspace{-0.3cm}
\begin{table} [H]
\centering
\caption{FOD Detection Rate For IoU $>$ 0.3}
\vspace{-0.3cm}
\begin{tabular}{cc}
\hline
Model & Detection Rate  \%\\
\hline
Entirely Convolutional & 75.3 \\
Latent ViT & None-Weak\\
Outer ViT & \textbf{82.7}\\
Skip Connections & None-Strong\\
\hline
\end{tabular}
\vspace{-0.3cm}
\label{tbl:ae}
\end{table}

\subsection{Classification}\label{subsec:classificationexperiments}
As mentioned in section~\ref{sec:related}, the FOD-A dataset provides a dataset of common FOD objects. The data is cropped to FOD-A's bounding boxes prior to training and validation to provide scenarios that are similar to the localization results.

\vspace{-0.3cm}
\begin{table} [H]
\centering
\caption{Classifier Comparison}
\vspace{-0.3cm}
\begin{tabular}{cc}
\hline
Model & Val Accuracy (FOD-A)\%\\
\hline
ResNet50V2 & 99.91\\
MobileNetV2 & 99.91\\
DenseNet169 & \textbf{99.94}\\
\hline
\end{tabular}
\vspace{-0.3cm}
\label{tbl:classifier}
\end{table}

The classifiers are trained with the cropped FOD-A dataset and utilized for classification. Accuracy results between classifiers are similar, with the highest validation result from DenseNet169 as shown in table~\ref{tbl:classifier}. Classification failed on a few cases where either the debris was too similar in color to the runway, the shadow of the debris seemed to be erroneously counted as part of the debris, or the object in the validation data was too dissimilar from the training data. Classification could likely be improved further with expansions to the FOD-A dataset to include and account for variables such as shadows at varying degrees as well as different styles of the same class of debris.

\vspace{-0.3cm}
\begin{table} [H]
\caption{Comparison With Image Based FOD Detection}
\centering
\begin{tabular}{cccc}
\hline
Method & Detection & Supervised & Environment  \\
 & Rate \% & & Specific \\
\hline
Ours & \textbf{82.7} & \textbf{No} & \textbf{No}\\
FOD-A SSD \cite{FOD-A} & 79.6 & Yes & No\\
FOD-A YOLO \cite{FOD-A} & 66.7 & Yes & No\\
Image Lookup \cite{FOD-Subtraction} & N/A & No & Yes\\
\hline
\end{tabular}
\label{tbl:related}
\end{table}

\subsection{Comparative Study}\label{sec:relatedcomparison} 

\begin{figure}[!h]
    \centering
    \vspace{-1.0em}
    \includegraphics[width=.9\linewidth]{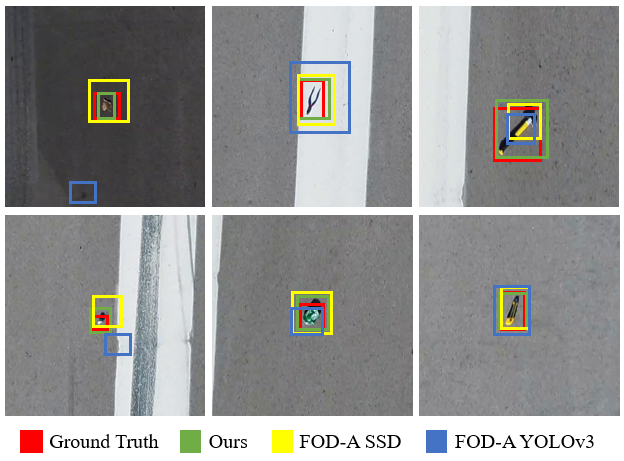}
    \vspace{-1.0em}
    \caption{A few visual comparison examples.}
    \vspace{-1.9em}
    \label{fig:visualexamples}
\end{figure}

We have studied the proposed method by comparing it against some state-of-the-art and related methods. 
As shown in table~\ref{tbl:related}, our method provides a stronger detection rate while being the only method that is not supervised and is not environment-specific. For an environment-specific method, the data has to be recollected for each airport implementation. Since our method is not supervised and is not environment-specific, it can accurately (supported by the top detection rate) detect FOD objects not included in training data, and can also generalize to new airports/environments. With respect to the image lookup method, our approach is not comparable because the former could only detect white or black objects, while our method can reasonably detect any non-pavement colored object.
Besides the quantitative summary, we present a few detection examples in figure~\ref{fig:visualexamples}, where we can observe that our proposed method shows better detection.

\textbf{}

\vspace{-0.75em}
\section{Conclusion}\label{sec:conclusion_discussion}
\vspace{-0.25em}

Although there are existing methods to detect FOD using technologies such as radar~\cite{faa-fact-sheet}, these approaches can be extremely expensive. Therefore, we provide a computer vision-based solution for FOD detection, and the dataset creation framework that can support this method. Computer vision can be significantly cheaper than the radar-based solutions, as the main requirements are a camera and some development time. There are also other image-based FOD detection methods, but they have weaknesses that may reduce their impact. The approach proposed in this paper solves these primary issues, such as reducing the data requirements and providing a method that can generalize to new objects.

The localization method in this paper may generalize to datatypes where the background-scenarios are known in advance. One research path following this paper could explore this generalization. Additionally, this paper discusses a learning block, a construct where a convolutional layer and ViT layer are easily interchangeable. This organizational construct streamlines experiments comparing the performance impact of interchanging convolutional and ViT layers in models. Additional work could explore scenarios where the ViT learning layers outperform the convolutional learning layers. 
Our team is also considering engineering solutions outside the computer vision domain, such as the addition of controlled floodlights in the UAS to further improve the original data under non-optimal lighting conditions.

\section{Resources}
The link to the dataset hosted on GitHub:

\href{https://github.com/FOD-UNOmaha/FODAnomalyData}{https://github.com/FOD-UNOmaha/FODAnomalyData}

\bibliographystyle{IEEEtran}
\bibliography{IEEEabrv,references}


\end{document}